\pdfoutput=1
\PassOptionsToPackage{table,xcdraw}{xcolor}

\documentclass[11pt]{article}

\usepackage[final]{acl} 

\usepackage{times}
\usepackage{latexsym}

\usepackage[T1]{fontenc}

\usepackage[utf8]{inputenc}

\usepackage{microtype}

\usepackage{inconsolata}

\usepackage{graphicx}

\usepackage{algorithm}
\usepackage{algorithmic}
\usepackage{booktabs}
\usepackage{array}
\usepackage{lscape}
\usepackage{paralist}
\usepackage{amsmath}
\usepackage{colortbl}

\usepackage{newfloat}
\usepackage{listings}
\usepackage{enumitem}
\usepackage{tablefootnote}

\title{Health Sentinel: An AI Pipeline For Real-time Disease Outbreak Detection}

\author{ \bf
Devesh Pant$^{1\dagger}$,
Rishi Raj Grandhe$^2$,
Jatin Agrawal$^2$,
Jushaan Singh Kalra$^2$, \\
\bf
Sudhir Kumar$^1$,
Saransh Khanna$^1$,
Vipin Samaria$^1$,
Mukul Paul$^1$, \\
\bf
Dr. Satish V Khalikar$^1$,
Vipin Garg$^1$, 
Dr. Himanshu Chauhan$^3$,
Dr. Pranay Verma$^3$, \\
\bf
Akhil VSSG$^2$,
Neha Khandelwal$^2$,
Soma S Dhavala$^2$, 
Minesh Mathew$^{1\dagger}$ \\  
$^1$Wadhwani AI, India, 
$^2$Work done while working at Wadhwani AI \\
$^3$National Centre for Disease Control, Government of India \\
$^\dagger$\texttt{\{devesh, minesh\}@wadhwaniai.org}
}

\begin{document}
\maketitle
\begin{abstract}
Early detection of disease outbreaks is crucial to ensure timely intervention by the health authorities. Due to the challenges associated with traditional indicator-based surveillance, monitoring informal sources such as online media has become increasingly popular. However, owing to the number of online articles getting published everyday, manual screening of the articles is impractical. To address this, we propose Health Sentinel. It is a multi-stage information extraction pipeline that uses a combination of ML and non-ML methods to extract events---structured information concerning disease outbreaks or other unusual health events---from online articles. The extracted events are made available to the Media Scanning and Verification Cell (MSVC) at the National Centre for Disease Control (NCDC), Delhi \footnote{\href{https://idsp.mohfw.gov.in/index4.php?lang=1&level=0&linkid=411&lid=3694}{Media Scanning and Verification Cell (MSVC)}} for analysis, interpretation and further dissemination to local agencies for timely intervention.
From April 2022 till date, Health Sentinel has processed over 300 million news articles and identified over 95,000  unique health events across India of which over 3,500 events were shortlisted by the public health experts at  NCDC as potential outbreaks.

\end{abstract}

%

\section{Introduction}
Disease surveillance is the continuous collection, analysis and interpretation of health data, particularly data concerning disease outbreaks and other unusual health events. It is essential that disease surveillance collects information in real time for timely interventions. Further, continuous disease surveillance allows us to monitor disease spread patterns and allocate resources more effectively, directing attention and funding to areas where they are most needed.

\begin{figure}[!hbt]
\centering
\includegraphics[width=0.95\columnwidth]{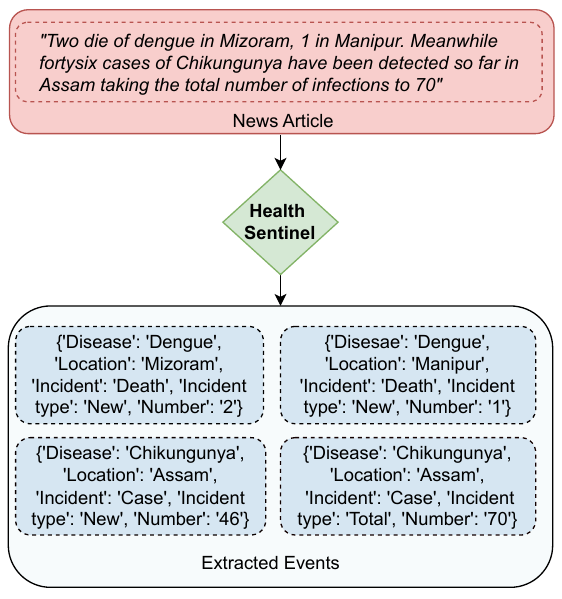}
\caption{Health Sentinel extracts structured information from online articles reporting unusual health events. The given example shows how our pipeline extracts multiple events from a single news article.}
\label{fig_teaser_figure}
\end{figure}

Traditional surveillance approaches follow a bottom-up approach wherein information is collected from health care workers, public health facilities and hospital networks. This approach is commonly referred to as `Indicator-based surveillance’.  It typically involves confirmed case reports, laboratory results, and clinical diagnoses. While indicator-based surveillance ensures that the data collected is mostly reliable, delay in reporting is often a concern. 
Further, weaker public health systems, and under-reporting  particularly in remote and rural areas  make indicator-based surveillance challenging~\cite{who_limitations_of_indicator_based}.

In contrast to the  indicator-based surveillance, event-based surveillance looks at multiple sources of information either formal or informal  such as print media reports,  online articles, and social media posts. This approach is designed to detect unusual health events quickly, providing early warnings for potential outbreaks. Owing to the nature of sources, data collected in this approach is likely to be noisy, redundant, and unstructured. Consequently, Information Retrieval (IR) and Natural Language Processing (NLP) techniques are increasingly being used in event-based surveillance systems to filter out irrelevant and redundant data to extract cohesive structured information.~\cite{communicable_diseases_surveillance_Review,animal_disease_thesis,eventepi_abbood,huff_2016_surveillance,padi_valentine}

According to the 2011 census, India has a population exceeding 1.2 billion~\cite{Census2011}. Most of the neglected tropical diseases are prevalent in India. Health threats triggered by climate change~\cite{lancet_climate_2023, lancet_climate_2024} is another significant concern.
Health governance in India is decentralized where  states hold primary responsibility for healthcare.  Traditional, indicator-based disease surveillance in such a setting demands a well coordinated system involving stakeholders belonging to different departments and different state governments. 

In this work, we present Health Sentinel, an information extraction pipeline, that feeds structured information concerning public health events to an event-based surveillance system in India. 
It extracts events related to $122$ human and animal diseases which were prioritized based on inputs from public health experts.

As shown in Figure \ref{fig:pipeline}, Health Sentinel follows a multi-stage process. 
It begins with data ingestion where  articles are periodically crawled from the web. Followed by this step, a binary text classifier filters out irrelevant articles. Next, all articles are translated to English.
Once an article is identified to carry information on one or more  unusual health events, we extract structured information from it. This is referred to as Event Extraction (EE)~\cite{event_extraction_survey} in Information Retrieval (IR) and Natural Language processing (NLP). In  our case, an ``event'' comprises,  i) Disease- the specific disease or ``others'' if the disease is not among a predefined list of $122$ diseases our health experts have curated, ii) Location  - the geographical area where the disease occurrence is reported, iii)  Incident - the nature of the event, such as case or death , iv) Incident type - whether the incident is New or Total (cumulative),  and  v) Number - the numerical value associated with the incident and its type (number of cases or number of fatalities). An example where 4 distinct events are extracted from a single article is shown in Figure~\ref{fig_teaser_figure}.
Followed by event extraction, similar  events  are clustered together to isolate unique event occurrences. The unique events are finally passed on to an expert for further review.

The highlights of Health Sentinel are listed as follows:
\begin{enumerate}[nosep]
    \item Health Sentinel, unlike most existing systems, scans the entire internet scouting for unusual health events.
    \item To the best of our knowledge, Health Sentinel is the first system that supports media scanning in multiple Indian languages. It supports 13 languages: English, Hindi, Telugu, Kannada, Gujarati, Tamil, Punjabi, Bengali, Marathi, Malayalam, Oriya, Assamese, and Urdu.
    \item We demonstrate that LLMs, including the recent open-source models perform better for event extraction  compared to the previously popular approaches like Named Entity Recognition (NER) and Question Answering (QA). 
    \item We propose a clustering logic that uses language model embeddings for text similarity and  DFS search on a graph built based on pairwise similarities and curated rules. 
    \item  Since its inception in April 2022, Health Sentinel has identified over 95,000 unique health events of which over 3,500 events were shortlisted by public health experts at NCDC.
    
\end{enumerate}
\section{Related Work}

Most ML-based disease surveillance approaches employ rule-based techniques along with classical ML models~\cite{disease_surveillance_ml_survey_Cabatuan_2020,disease_surveillance_ml_survey_Zeng_2021} .
MediSys (J et al., 2010) and ProMED (Yu and Madoff, 2004) are the two most popular disease surveillance systems. MediSys uses pattern-matching techniques to extract events from articles which leads to many false positives. Since it is rule-based, extending it to other languages is non-trivial. On the other hand, in ProMED, filtering relevant information and further analysis is mostly performed by involving humans.
In recent times, new tools like GRITS~\cite{huff_2016_surveillance} , EventEpi~\cite{eventepi_abbood} and Padi 3.0~\cite{padi_valentine} have been developed for disease surveillance. Although these systems use ML for tasks like classification and clustering, none of them exploit recent advances in event extraction using deep learning techniques and LLMs.

\noindent EE, or extracting structured information from unstructured text, is a well studied problem in IR and NLP. Deep learning-based NER and QA models have extensively been used for EE tasks~\cite{event_extraction_survey}. However, LLMs are increasingly being used for EE tasks in zero-shot, few-shot and finetuned settings~\cite{event_extraction_2024_survey}. LLMs have demonstrated impressive results in information extraction tasks using few-shot approach without requiring task-specific fine-tuning~\cite{llm_fewshot}. These results have further been validated by studies exploring use of OpenAI models like GPT-3.5 and GPT-4\footnote{\url{https://platform.openai.com/docs/models}} for EE in various scenarios~\cite{llm_chatie_2024,llm_chat_extract_materials,llm_gao_chatgpt_ee}. 
Similar to these works, we use LLMs for few-shot EE.
Dagdelen et al. demonstrate that  GPT-3 and Llama-2~\cite{llama2} can be finetuned for extracting structured information from scientific text. We have not explored LLM finetuning for EE for disease surveillance owing to the lack of training data. Harrod et al.~\cite{nlp4pi_disease_map} use LLMs for extraction of structured epidemiological data from documents and geotagging each record. Their work is similar to ours as they extract structured disease related information and use LLMs for the same. However, their work focuses on extracting information concerning Rift Valley Fever (RVF) alone. Secondly, the objective of their work is not disease surveillance but creation of a structured epidemiological dataset for RVF from past documents---PDFs of research articles and other documents concerning RVF. In contrast, our work uses LLMs for event extraction from web articles for real time surveillance of 100+ diseases.

\section{Method}

\begin{figure*}[!hbt]
\centering
\includegraphics[width=1.0\textwidth]{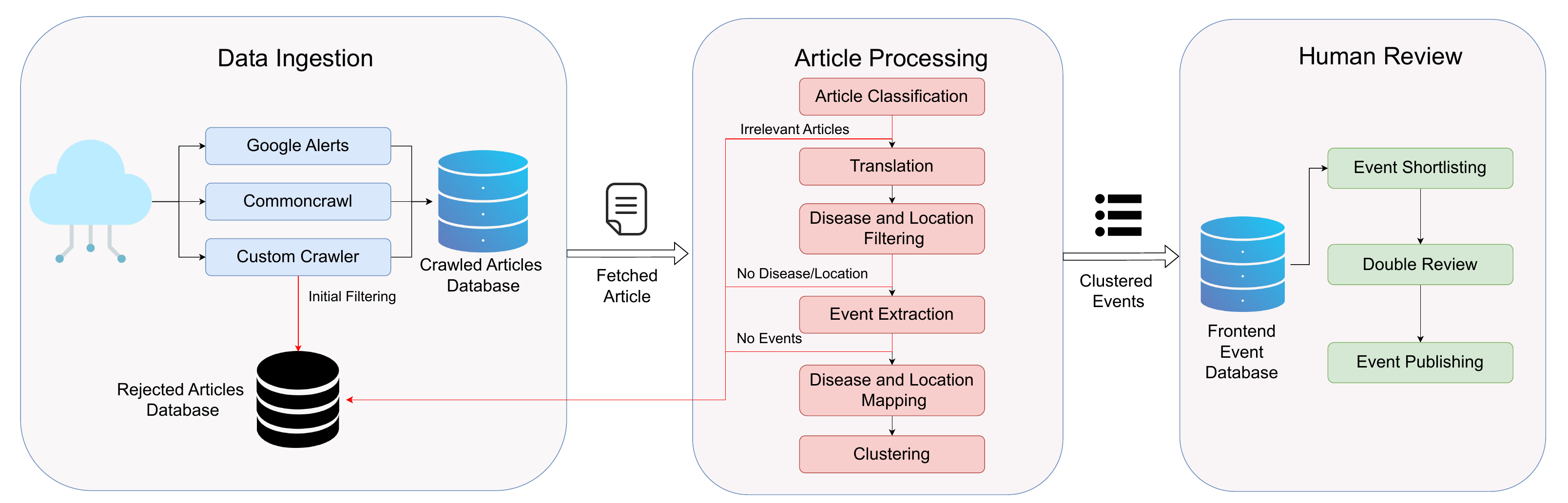}
\caption{System Overview of Health Sentinel. Health Sentinel combines rule based and ML techniques alongside a human-in-the-loop system to ensure a high level of reliance and efficiency. Its data ingestion pipeline continuously collects news articles from the web and stores them in a database. The article processing pipeline retrieves these articles, filters out irrelevant data, and extracts health events. The extracted events are then sent for expert review before publication for ground-level action. }
\label{fig:pipeline}
\end{figure*}

In the following sections, we present details of each stage in the Health Sentinel system. The overall flow can be referred to in Figure \ref{fig:pipeline}. 
\subsection{Data Ingestion}
For a real-time system, it is essential to continuously monitor the web for newly added articles. Health Sentinel achieves this using three services: Common Crawl\footnote{\url{https://github.com/commoncrawl/news-crawl/}}, Google Alerts\footnote{\url{https://www.google.com/alerts}} and custom crawlers.
We use Common Crawl's news database to fetch the latest published articles every few hours. We configured Google Alerts using keywords in 12 Indic languages and English. The keywords for Google Alerts were selected by public health experts based on the 122 disease/health events that we are interested in monitoring. We have additionally designed custom crawlers for a few news websites that are not covered by the former two services.

From the URLs of the news articles collected through these sources,  `title' and `description' tags are extracted. These tags provide concise information about the webpage content, such as headline of the news and a summary of the article body. Given the extremely wide scope of HTML-based webpages, it is infeasible to  effectively extract the relevant content of the webpage from its body while isolating noisy information like advertisements. Therefore, for further processing, the text used from an article is the concatenation of the `title' and `description' fields.

The articles undergo a rule-based filtering based on three criteria: a domain blocklist, recency (only the most recently published articles are retained), and language. The block-list contains domain names of non-Indian news websites, allowing us to filter out more than $90\%$ of irrelevant articles that cover news outside India.
The source language of an article is identified using langid ~\cite{paper_langid}, and only articles in the 13 supported languages are retained.

\subsection{Article Classification}
\label{sec:article_classification}
A substantial portion (nearly $87\%$) of the articles at this stage are irrelevant to Health Sentinel as they do not carry any health events-related information. A  keyword-spotting mechanism fails to filter out irrelevant ones as it cannot take the article's context into account. For instance, ``What is Dengue? 10 ways to stay safe this monsoon'' is an article related to human health but doesn't contain any actionable event information. Therefore, we train a binary classifier to discard  irrelevant articles and effectively reduce the throughput for  stages downstream, particularly translation and event extraction. 
To develop this classifier, we finetuned  multiple Transformer-based, encoder-only  (BERT-like) models that had  been pretrained for language modeling tasks.
For English, we experimented with six different models. For the  Indic languages, we tried out  four models selected based on their general performance on these languages. List of all the models we  tried out is given in Section \ref{supp:art_cls_subsec} in the Appendix. The best-performing model for each of the 13 supported languages was selected based on the models' validation set performance.

\subsection{Translation}
Once relevant articles are identified, we translate them into English. This is necessary because most of the ML models including LLMs used in the subsequent stages of the pipeline perform better in English compared to low-resource Indian languages~\cite{li2024quantifyingmultilingualperformancelarge}. 
While paid APIs such as Google Translate\footnote{\url{https://cloud.google.com/translate}} and Microsoft Azure Translate\footnote{\href{https://azure.microsoft.com/en-us/products/cognitive-services/translator}{azure translator}} have long been preferred for low-resource languages, recent open-source models including \texttt{IndicTrans}~\cite{indic_trans_v1} and \texttt{IndicTrans2}~\cite{gala2023indictrans}  perform on par or even better than these APIs for many Indic languages ~\cite{gala2023indictrans}. 
Due to the superior performance on most of the Indic languages translation benchmarks, we use \texttt{IndicTrans2} in our pipeline for translating articles into English. 

\subsection{Disease and Location based Filtering}
\label{sec:disease_and_location_filtering}
Though our article classifier significantly reduces the number of irrelevant articles, some still pass through. Moreover, despite domain-block-listing, a large share of articles collected from the web discuss health events outside India. 
To address this, we implement additional filters to ensure that each article mentions both a disease related to humans, animals, or plants and an Indian location.
To identify diseases, we use an ensemble approach combining keyword-spotting with a disease NER model. The keywords include scientific names and common synonyms used in the media for diseases relevant to our system, curated by experts. 
For the disease NER, we use the open-source BioBERT\footnote{\href{https://huggingface.co/alvaroalon2/biobert_diseases_ner}{biobert disease ner}}.
For location identification, we construct an exhaustive list of Indian locations, including names of states, districts, sub-districts, and their synonyms used in the media. This list is provided to an NER model\footnote{\href{https://github.com/hellohaptik/chatbot_ner}{chatbot ner}}, which identifies the locations mentioned in the article. 
Any article that does not mention both a relevant disease and an Indian location is discarded.

\subsection{Event Extraction}
\label{sec:event_extraction}
While developing Health Sentinel, we explored two  approaches for event extraction. The first approach uses a combination of QA and NLI, while the second one uses LLMs.
\subsubsection{Event Extraction using QA and NLI}
\label{event_extraction_qa_nli}
In this approach, we use previously extracted location and disease data (see Section~\ref{sec:disease_and_location_filtering}) for retrieving remaining entities---Incident, Incident type and Number.\newline
\noindent\textbf{Numbered Events Extraction: } News articles reporting health events often include numerical information about cases or deaths (see the example in Figure \ref{fig_teaser_figure}). We use a QA model to extract such numbers by asking structured questions such as: \textit{``How many {new cases} of [Disease] are there in [Location]?''} or \textit{``How many {total deaths} due to [Disease] were reported in [Location]?''}. If the model provides an answer, the entities used in the question, along with the extracted numerical value form an event. To ensure comprehensive extraction, we have carefully designed a diverse set of questions to account for variations in how the information may be presented in articles (see Table \ref{tab:questions} in the Appendix). These templates cover different combinations of Incident (cases vs. deaths) and Incident type (new vs. total), while disease and location are dynamically inserted. 
For this task, we use   \texttt{deepset-roberta-large-squad2}\footnote{\href{https://huggingface.co/deepset/roberta-large-squad2}{roberta-large-squad2}}, an off-the-shelf extractive QA model.

\noindent\textbf{Numberless Events Extraction: }Some articles discuss important health information without providing numerical data. For example, a statement such as ``Dengue is on the rise in Karnataka'' highlights a significant health concern but lacks explicit numbers, while ``Monkeypox: No need to be afraid, says Kerala Health Minister'' contains disease information but no actionable event. The absence of numerical information makes it
challenging to differentiate between actionable health events
and general health information. 
To handle such cases, we use NLI. Hypotheses such as \textit{``{Cases} of [Disease] have risen in [Location]''} or \textit{``People are {dying} of [Disease] in [Location]''} are generated, and the article text is provided as the premise to the NLI model. If the model determines that the premise entails the hypothesis, the corresponding combination of disease, location, and incident is considered as an event. For this task, we use off-the-shelf NLI model \texttt{microsoft-deberta-large-nli}\footnote{\href{https://huggingface.co/microsoft/deberta-large-mnli}{deberta-large-mnli}}.

\subsubsection{Event Extraction Using LLMs}

To implement this approach, we designed a system prompt \( P \) that assigns the task of event extraction to the LLM. The prompt includes descriptions of each entity that constitutes an event   and guides the model's response generation through few-shot examples \( \{E\} \). 

Formally, the LLM takes an article \( A \) as input and generates a structured JSON response:
\[
   \mathcal{E} = \text{LLM}(A, P, \{E\}) = \{ e_1, e_2, \dots, e_n \}
\]
where each extracted event \( e_i \) is a dictionary containing the set entities---Disease, Location, Incident, Incident type and Number---that forms the event.
\[
e_i = \{ (k_1, v_1), (k_2, v_2), \dots, (k_m, v_m) \}
\]
where \( k_j \) represents an entity and \( v_j \) is its corresponding value extracted from \( A \).
We also leverage LLMs' capability to filter out irrelevant content that may have bypassed earlier filtering at the article classification stage (see Section~\ref{sec:article_classification} ). The prompt explicitly distinguishes between general health information and actionable health events, instructing the model to focus solely on the latter. An example prompt is shown in Table \ref{tab:llm_prompt} in the Appendix. 
Articles with no events extracted by the LLM are re-processed using another prompt, serving as a double-check for the LLM's extraction. This prompt focuses on identifying events without numerical information, similar to the NLI approach described in the previous method. \newline We experimented with various prompt designs and selected the most effective one based on both quantitative and qualitative evaluations. We have experimented with both proprietary LLMs and open-source ones. Table \ref{tab:eval_end_to_end} can be referred to for the list of LLMs we have tried out for the event extraction. 

\subsection{Mapping of Disease and Location}
This stage  ensures that the extracted disease and location names align with standardized disease and location names used by the Media Scanning and Verification Cell. For disease mapping, we first use a curated dictionary that maps common synonyms and media terms to standardized disease names. If an extracted disease does not get mapped this way, we use an LLM to map it to the nearest standard name. For location mapping, we employ a hierarchical dictionary to assign extracted locations to administrative levels such as states, districts, and sub-districts. For any extracted location that fail to get mapped using the above approach, we prompt an LLM to map the location to an Indian state (see Section\ref{sec:disease_mapping} for more details).

\subsection{Clustering}
A health event is often reported by multiple media outlets and other online sources. Since the previous stages in our pipeline do not check if an extracted event is a duplicate of another, we use a clustering mechanism at the end to find clusters of unique events.
This stage uses a combination of ML techniques and  rules to isolate unique health events. Articles are only clustered at a day-level to maintain consistency and ease of use. The clustering involves the following steps.
\begin{enumerate}[nosep]
    \item A pretrained sentence transformer, \texttt{paraphrase-distilroberta-base-v2}\footnote{\href{https://huggingface.co/sentence-transformers/paraphrase-distilroberta-base-v2}{paraphrase-distilroberta-base-v2}}., is used to create an embedding of the article associated with an event.
    \item Cosine similarity is calculated for every pair of article embeddings to generate a 2D similarity matrix for each pair of events.
    \item A rule set is used to analyze the extracted event information for every pair of events to determine the threshold to apply on the similarity score. Using the threshold, each similarity score is  set 0 or 1. This creates a 2D match matrix with 1's and 0's.
    \item A  Depth First Search (DFS) is performed on the match matrix to get  all the disjoint graphs. Each disjoint graph is treated as a cluster.
    \item We run a conflict check on each cluster and further break it down if it has any events with conflicting information.
\end{enumerate}
Further details on the clustering are given in Section \ref{sec:clustering} in the Appendix.

\subsection{Human-In-the-Loop}
Before any action is taken on the extracted health events, public health experts at NCDC review them using on-ground epidemiological indicators.

\section{Experiments and Results}
\label{sec:experiments}

\subsection{Datasets} 
\subsubsection{Article Classifier Dataset} 
    We collected $34,527$ English articles sourced from the internet and manually labeled  them as relevant or irrelevant. This dataset consists of $7,374$ articles in the positive class and $27,153$ articles in the negative class. This dataset was further split into training, validation, and test sets while ensuring  that the test set contains a representative range of diseases to validate the classifier's performance across different scenarios. 
    In order to train the classifier for other languages, the English dataset is translated into other 12 Indic languages using the 
    \texttt{IndicTrans2} model.
    
\subsubsection{End-to-End Evaluation Dataset} Articles in this dataset were sourced from news articles captured by a human-based media disease surveillance system for a period before Health Sentinel's deployment. Out of $1005$ articles in the dataset, $610$ contain  events (relevant articles), and $395$ have no events (irrelevant articles). The dataset contains $71$ unique diseases across more than $250$ unique locations in India.

\subsubsection{Clustering Evaluation Dataset} The dataset has $869$ events spread across $7$ different dates with $503$ clusters that are clustered on a per-day basis as shown in Table \ref{tab:combined_cluster}. The ideal cluster compositions were annotated by health experts.



\begin{table*}[!hbt]
\centering
\small
\begin{tabular}{lcccccc}
\toprule
\textbf{Language} & \textbf{Model} & \textbf{Accuracy} & \textbf{Precision} & \textbf{Recall} & \textbf{F1-Score} & \textbf{AUC-ROC} \\
\midrule
English & roberta-base & 0.99 & 0.99 & 0.96 & 0.97 & 0.98\\
Hindi & google/muril-base-cased & 0.98 & 0.96 & 0.97 & 0.96 & 0.98\\
Telugu & xlm-roberta-base & 0.98 & 0.97 & 0.95 & 0.96 & 0.97\\
Kannada & google/muril-base-cased & 0.98 & 0.98 & 0.96 & 0.97 & 0.98\\
Gujarati & google/muril-base-cased & 0.98 & 0.96 & 0.96 & 0.96 & 0.98\\
Tamil & google/muril-base-cased & 0.99 & 0.97 & 0.96 & 0.97 & 0.98\\
Punjabi & xlm-roberta-base & 0.98 & 0.96 & 0.95 & 0.96 & 0.97\\
Bengali & xlm-roberta-base & 0.98 & 0.97 & 0.95 & 0.96 & 0.97\\
Marathi & xlm-roberta-base & 0.98 & 0.97 & 0.96 & 0.96 & 0.97\\
Malayalam & google/muril-base-cased & 0.98 & 0.95 & 0.95 & 0.95 & 0.97\\
Oriya & xlm-roberta-base & 0.98 & 0.96 & 0.94 & 0.95 & 0.97\\
Assamese & google/muril-base-cased & 0.98 & 0.95 & 0.95 & 0.95 & 0.97\\
Urdu & xlm-roberta-base & 0.98 & 0.96 & 0.96 & 0.96 & 0.97\\
\bottomrule
\end{tabular}
\caption{Performance of the best classification models for each language, evaluated on the respective test sets. All models are downloaded from \url{https://huggingface.co/models} and finetuned on the respective training data.}
\label{tab:main_eval_artclf}
\end{table*}

\subsection{Results}
\subsubsection{Article Classifier}
\label{subsubsec_artclf_results}
After experimenting with multiple BERT-like models, we selected the best model for each language based on recall. The results are presented in Table \ref{tab:main_eval_artclf}. For English, the \texttt{roberta-base}~\cite{liu2019roberta} model performed the best. For other languages, \texttt{google/muril-base-cased} and \texttt{xlm-roberta-base} yielded the best results.
We observe that these classifiers isolate  non-health related articles as irrelevant with near perfect accuracy. However, they tend to struggle with health-related articles  that do not contain any health events. For instance, ``Exclusive: Monsoon can host a buffet of illnesses. Doctor reveals secrets to guarding against seasonal infections, allergies'' was considered  relevant by the article classifier. Overall, all selected models achieve a recall and F1-score of approximately 96\%, making them highly effective as an initial filter for irrelevant articles.

\begin{table*}[!hbt]
\centering
\footnotesize
\begin{tabular}{lccccccccccc}
\toprule
\textbf{Model} & \multicolumn{3}{c}{\textbf{Event Extraction}} & \textbf{Exact Match} & \textbf{Detection} & \multicolumn{3}{c}{\textbf{Disease Extraction}} & \multicolumn{3}{c}{\textbf{Location Extraction}} \\
\cmidrule(lr){2-4} \cmidrule(lr){7-9} \cmidrule(lr){10-12}
& \textbf{P} & \textbf{R} & \textbf{F1} & \textbf{Accuracy} & \textbf{Rate} & \textbf{P} & \textbf{R} & \textbf{F1} & \textbf{P} & \textbf{R} & \textbf{F1} \\
\midrule
QA+NLI based Pipeline & 0.41 & 0.40 & 0.40 & 0.37 & 0.70 & 0.55 & 0.52 & 0.54 & 0.52 & 0.49 & 0.50 \\
\midrule
Llama3.1-8b\footnote{\url{https://ollama.com/library/llama3:8b-instruct-q4_0}} & 0.50 & 0.50 & 0.50 & 0.43 & 0.95 & 0.77 & 0.79 & 0.78 & 0.68 & 0.70 & 0.69 \\
Gemma2-9b\footnote{\url{https://ollama.com/library/gemma2:9b-instruct-q4_0}} & 0.54 & 0.50 & 0.52 & 0.45 & \textbf{0.96} & \textbf{0.84} & 0.78 & 0.81 & 0.77 & 0.71 & 0.74 \\
GPT3.5-Turbo & 0.62 & 0.61 & 0.61 & 0.54 & 0.95 & 0.81 & 0.79 & 0.80 & 0.78 & 0.76 & 0.77 \\
GPT-4o-Mini& \textbf{0.70} & \textbf{0.67} & \textbf{0.68} & \textbf{0.61} & 0.92 & 0.83 & \textbf{0.80} & \textbf{0.81}& \textbf{0.81} &\textbf{0.77} & \textbf{0.79} \\
\bottomrule
\end{tabular}
\caption{Performance comparison of end-to-end event extraction using different models, showing results across multiple metrics--- precision (P), recall (R), and F1-score at the event-level, and for individual entities--- disease and location. LLM-based pipelines achieve significantly better results compared to the QA and NLI-based methods, with GPT-4o-Mini performing best overall.}

\label{tab:eval_end_to_end}
\end{table*}

\begin{table*}[h!]
\centering
\footnotesize
\renewcommand{\arraystretch}{1.3} 
\setlength{\tabcolsep}{5pt}       
\resizebox{\textwidth}{!}{ 

\begin{tabular}{|>{\centering\arraybackslash\columncolor[HTML]{FFF3E0}}p{1cm}|
                >{\centering\arraybackslash\columncolor[HTML]{E3F2FD}}p{6cm}|
                >{\centering\arraybackslash\columncolor[HTML]{FFE5E5}}p{5cm}|
                >{\centering\arraybackslash\columncolor[HTML]{E8F5E9}}p{5cm}|}
\hline
\rowcolor[HTML]{ECEFF1}
\textbf{\#} & \textbf{Article} & \textbf{QA+NLI Pipeline} & \textbf{GPT-4o-mini Pipeline} \\ \hline

1 & Mysterious Disease In AP's Eluru Claims 1 Life, 347 Falls Ill, Samples Sent To Delhi. & 
\{`Disease': `Falls', `Location': `Eluru', `Incident': `death', `Incident type': `total', `Number': `347'\} & 
[\{`Disease': `Mysterious Disease', `Location': `Eluru', `Incident': `case', `Incident type': `new', `Number': `347'\}, \{`Disease': `Mysterious Disease', `Location': `Eluru', `Incident': `death', `Incident type': `new', `Number': `1'\}]\\ \hline

2 & Corona turmoil in North Korea.. 21 people died of fever. North Korea | North Korea (North Korea) is trembling with fever. & 
\{`Disease': `Corona', `Location': `Korea', `Incident': `death', `Incident type': `new', `Number': `21'\} & 
[]  \\ \hline

3 & In Himachal, 535 people admitted to hospital after drinking contaminated water. & 
[] & 
\{`Disease': `Food poisoning infection', `Location': `Himachal', `Incident': `case', `Incident type': `new', `Number': `535'\} \\ \hline

4 & Mancherial brothers' death: Two brothers passed away within hours.. knowing that the younger brother had died of a heart attack.. the elder brother went there and got a heart attack. & 
\{`Disease': `Cardiac arrest', `Location': `Mancherial', `Incident': `death', `Incident type': `\_', `Number': `\_'\} & 
[] \\ \hline

\end{tabular}}
\caption{Qualitative comparison of event extraction by GPT-4o-Mini and QA+NLI pipelines. In example 1, the LLM-based pipeline identifies a disease missed by QA+NLI. In Example 2, it filters out an irrelevant international event mistakenly extracted by QA+NLI. Example 3 shows the LLM capturing an illness caused by contaminated water, which QA+NLI misses. In the final example, it excludes the article lacking an infectious disease component, unlike QA+NLI. Overall,  LLM's inherent knowledge enables more accurate event extraction and contextual filtering of articles.}
\label{tab:qualitative_results}
\end{table*}

\subsubsection{Event Extraction}
Results of event extraction are shown in Table \ref{tab:eval_end_to_end}. We report event-level precision, recall, and F1-score to evaluate overall performance. We also evaluate location and disease extraction separately to highlight entity-specific performance. Additionally, exact match accuracy measures how closely the extracted events resemble the ground truth, while detection rate reflects the model's ability to extract at least one event in relevant articles.

As shown in Table \ref{tab:eval_end_to_end}, LLMs surpass traditional NER methods in extracting disease and location information. They effectively filter out irrelevant articles, such as those related to injuries, accidents, and general health information. The qualitative results are shown in Table \ref{tab:qualitative_results}. 
Among the tested LLMs, proprietary models outperform open-source ones. The GPT-4o-Mini model achieves the best overall results. However,  Llama3.1-8b and Gemma2-9b  show competitive performance.
Exact match accuracy is around $60\%$ even for the best-performing LLM. We observe that the models struggle to extract all events when multiple events are present in an article.
Additionally, LLMs sometimes misinterpret `new cases' as `total cases,' resulting in errors. Nevertheless, the system maintains high detection rate, ensuring that most relevant articles are captured for the human review stage.

\subsubsection{Clustering}
To quantitatively evaluate the quality of the clusters formed, we employ three key metrics, Adjusted Rand Index(ARI)~\cite{Hubert1985}, Normalized Mutual Information(NMI)~\cite{Strehl2002}, and V-Measure~\cite{Rosenberg2007} (explained in Section \ref{clustering_metrics} in the Appendix). Clustering performance is reported in Table \ref{tab:combined_cluster}. 
For the evaluation metrics, higher values indicate strong agreement between the generated clusters and the ground truth.




\begin{table}[t]
    \centering
    \footnotesize
    \setlength{\tabcolsep}{3pt} 
    \begin{tabular}{l p{12mm} p{12mm} p{10mm} p{10mm} p{10mm}}
    \toprule
    \textbf{Date} & \textbf{Data \newline Points} & \textbf{Clusters} & \textbf{ARI} & \textbf{NMI} & \textbf{V-Measure} \\
    \midrule
    05/24/24 & 91 & 72 & 0.94 & 0.99 & 0.99 \\
    05/25/24 & 63 & 55 & 0.89 & 0.99 & 0.99 \\
    05/26/24 & 64 & 46 & 0.99 & 1.00 & 1.00 \\
    06/09/24 & 107 & 85 & 0.84 & 0.99 & 0.99 \\
    06/10/24 & 81 & 65 & 0.91 & 0.99 & 0.99 \\
    06/11/24 & 103 & 79 & 0.79 & 0.99 & 0.99 \\
    06/21/24 & 360 & 101 & 0.84 & 0.94 & 0.94 \\
    \midrule
    \textbf{Avg.} & -- & -- & \textbf{0.89} & \textbf{0.98} & \textbf{0.98} \\
    \bottomrule
    \end{tabular}
    \caption{Per-day clustering dataset statistics and performance. Clusters are evaluated using Adjusted Rand Index (ARI), Normalized Mutual Information (NMI), and V-Measure.}
    \label{tab:combined_cluster}
\end{table}


\section{Deployment and Impact}
Health Sentinel was launched in April 2022  with support for  English and Hindi. Over the  two years, it has expanded the support to $11$ additional languages. The system has  undergone multiple upgrades, particularly in the event extraction module, to integrate the latest deep learning models and LLMs. To date, it has processed over 300 million articles from over 500,000 unique domains and has identified over 95,000 unusual unique health events. On a daily basis, Health Sentinel processes around 375,000 news articles  and identify around 150 unusual unique health events. Since deployment, over 3,500 events have been  shortlisted by the health experts at NCDC. Notably, only a small percentage of detected events were  shortlisted by the human experts. This can be attributed to the following: i)
duplicates that are not clustered correctly, ii)  common outbreaks that occur during expected seasons are often not shortlisted by the health authority, and  iii) if the disease is endemic in the location, such events are not shortlisted.

\noindent To better understand the impact of Health Sentinel, we compare it with the human-based surveillance. We observed the following, i) number of published events saw a $150\%$ increase compared to the previous years where only human-based surveillance existed ii) in 2024, $96\%$ of the health events published by the surveillance system were extracted by the Health Sentinel (only $4\%$ were found by manual scanning of the media), and iii) the number of media sources covered has grown exponentially because of automated media scanning and multilingual support.

\section{Limitations and Future Work}
In this section, we discuss known limitations of our system. Due to the lack of robust pre-trained models for Indic languages for event extraction and text embeddings, we need to translate non-English articles to English following the article classification stage. We have observed that named entities such as disease names and location names are sometimes mistranslated, leading to a lower performance for articles sourced from these languages. We experimented with fine tuning \texttt{IndicTrans2}~\cite{gala2023indictrans} framework with an emphasis on correctly translating or transliterating named entities as appropriate. While this improved translation of the entities, it led to a decline in overall performance of the model. 

Another known limitation is the lack of full context for the event extraction step, since we read only title and description of an article. Reading the full body of online articles is practically challenging since  the body text is almost always clubbed with advertisements and other unrelated content and the format varies from page to page. Currently we are developing custom HTML source parsers for selected websites so that full body of the articles from these websites can be read.


\section{Conclusion}
In this paper, we presented Health Sentinel for automating media-based disease surveillance in a multilingual setting. Health Sentinel works by leveraging the capabilities of different sequentially connected Machine Learning models to maintain an optimal level of latency while maximizing the ability to identify unusual health events. Health Sentinel has demonstrated promising results across multiple evaluation metrics and has greatly increased the capability of disease surveillance in India.


\section{Ethical and Societal Implications}
Relying on online content for disease surveillance presents several challenges that can impact the reliability of the information. News sources may introduce risks of misinformation or sensationalism in reporting. Moreover, the system might inherit biases in regional coverage or language representation, potentially leading to uneven event detection.

To address these challenges, we conducted a detailed assessment of online news sources across different regions and languages in the country and implemented the following measures:

\noindent \textbf{1. Keyword expansion for Indic languages.} We curated a list of 7,000 disease-related keywords for Google Alerts by translating and transliterating disease names into Indic languages. This ensures early detection from regional news sources, even before these events appear on national news websites.\newline
\noindent \textbf{2. Local news coverage.} We identified regional news websites that are often overlooked by platforms like Common Crawl or Google News. To address this, we developed a custom crawler that manually collects articles from these sources, improving regional representation.

\noindent \textbf{3. Source filtering and periodic reviews. }To handle the issue of misinformation, we maintain a list of unreliable  news sources. News sources flagged as unreliable are periodically evaluated and blacklisted from entering the system in the future. Additionally, the clustering feature in our pipeline helps group similar news articles about the same event. This allows human reviewers to cross-check information from multiple sources, identify fake news and mitigate the impact of exaggerated or inaccurate information. 

Despite these safeguards, some misinformation may still slip through the system. Additionally, using LLMs for event extraction can introduce noise due to hallucinations. Therefore, before the extracted information is published for field use, it must undergo a review by the health experts.

\bibliography{acl_latex}

@article{lancet_climate_2023,
  title={The 2023 report of the Lancet Countdown on health and climate change: the imperative for a health-centred response in a world facing irreversible harms},
  author={Romanello, Marina and Di Napoli, Claudia and Green, Carole and Kennard, Harry and Lampard, Pete and Scamman, Daniel and Walawender, Maria and Ali, Zakari and Ameli, Nadia and Ayeb-Karlsson, Sonja and others},
  journal={The Lancet},
  volume={402},
  year={2023}
}

@article{lancet_climate_2024,
  title={The 2024 report of the Lancet Countdown on health and climate change: facing record-breaking threats from delayed action},
  author={Romanello, Marina and Walawender, Maria and Hsu, Shih-Che and Moskeland, Annalyse and Palmeiro-Silva, Yasna and Scamman, Daniel and Ali, Zakari and Ameli, Nadia and Angelova, Denitsa and Ayeb-Karlsson, Sonja and others},
  journal={The Lancet},
  volume={404},
  year={2024}
}

@inproceedings{paper_langid,
author = {Lui, Marco and Baldwin, Timothy},
title = {langid.py: an off-the-shelf language identification tool},
year = {2012},
publisher = {Association for Computational Linguistics},
address = {USA},
booktitle = {Proceedings of the ACL 2012 System Demonstrations},
pages = {25–30},
numpages = {6},
location = {Jeju Island, Korea},
series = {ACL '12}
}

@article{indic_trans_v1,
    author = {Ramesh, Gowtham and Doddapaneni, Sumanth and Bheemaraj, Aravinth and Jobanputra, Mayank and AK, Raghavan and Sharma, Ajitesh and Sahoo, Sujit and Diddee, Harshita and J, Mahalakshmi and Kakwani, Divyanshu and Kumar, Navneet and Pradeep, Aswin and Nagaraj, Srihari and Deepak, Kumar and Raghavan, Vivek and Kunchukuttan, Anoop and Kumar, Pratyush and Khapra, Mitesh Shantadevi},
    title = "{Samanantar: The Largest Publicly Available Parallel Corpora Collection for 11 Indic Languages}",
    journal = {Transactions of the Association for Computational Linguistics},
    volume = {10},
    pages = {145-162},
    year = {2022},
    month = {02},
    issn = {2307-387X},
    doi = {10.1162/tacl_a_00452},
    url = {https://doi.org/10.1162/tacl\_a\_00452},
    eprint = {https://direct.mit.edu/tacl/article-pdf/doi/10.1162/tacl\_a\_00452/1987010/tacl\_a\_00452.pdf},
}

@article{gala2023indictrans,
title={IndicTrans2: Towards High-Quality and Accessible Machine Translation Models for all 22 Scheduled Indian Languages},
author={Jay Gala and Pranjal A Chitale and A K Raghavan and Varun Gumma and Sumanth Doddapaneni and Aswanth Kumar M and Janki Atul Nawale and Anupama Sujatha and Ratish Puduppully and Vivek Raghavan and Pratyush Kumar and Mitesh M Khapra and Raj Dabre and Anoop Kunchukuttan},
journal={Transactions on Machine Learning Research},
issn={2835-8856},
year={2023},
url={https://openreview.net/forum?id=vfT4YuzAYA},
note={}
}

@misc{li2024quantifyingmultilingualperformancelarge,
      title={Quantifying Multilingual Performance of Large Language Models Across Languages}, 
      author={Zihao Li and Yucheng Shi and Zirui Liu and Fan Yang and Ali Payani and Ninghao Liu and Mengnan Du},
      year={2024},
      eprint={2404.11553},
      archivePrefix={arXiv},
      primaryClass={cs.CL},
      url={https://arxiv.org/abs/2404.11553}, 
}

@misc{khanuja2021murilmultilingualrepresentationsindian,
      title={MuRIL: Multilingual Representations for Indian Languages}, 
      author={Simran Khanuja and Diksha Bansal and Sarvesh Mehtani and Savya Khosla and Atreyee Dey and Balaji Gopalan and Dilip Kumar Margam and Pooja Aggarwal and Rajiv Teja Nagipogu and Shachi Dave and Shruti Gupta and Subhash Chandra Bose Gali and Vish Subramanian and Partha Talukdar},
      year={2021},
      eprint={2103.10730},
      archivePrefix={arXiv},
      primaryClass={cs.CL},
      url={https://arxiv.org/abs/2103.10730}, 
}

@mis{who_limitations_of_indicator_based,
  title={A guide to establishing event-based surveillance},
  author={WHO and others},
  year={2008},
  publisher={WHO Regional Office for the Western Pacific},
     howpublished = {\url{https://iris.who.int/bitstream/handle/10665/207737/9789290613213_eng.pdf}},
note = "Accessed on 15 August 2024"
}

@misc{census2011,
  title = {Census 2011},
  howpublished = {\url{https://censusindia.gov.in/2011-Common/CensusData2011.Html}},
author = "Office of the Registrar general and census commissioner of India ",
note = "Accessed on 26 March 2025"
}

@phdthesis{animal_disease_thesis,
  TITLE = {{Extraction and combination of epidemiological information from informal sources for animal infectious diseases surveillance}},
  AUTHOR = {Valentin, Sarah},
  URL = {https://theses.hal.science/tel-03174891},
  SCHOOL = {{Universit{\'e} Montpellier}},
  YEAR = {2020}
}

@article{communicable_diseases_surveillance_Review,
  title={Surveillance of communicable diseases using social media: A systematic review},
  author={Pilipiec, Patrick and Samsten, Isak and Bota, Andr{\'a}s},
  journal={PLoS One},
  volume={18},
  number={2},
  year={2023}
}

@article{huff_2016_surveillance,
  title={Evaluation and verification of the global rapid identification of threats system for infectious diseases in textual data sources},
  author={Huff, Andrew G and Breit, Nathan and Allen, Toph and Whiting, Karissa and Kiley, Christopher},
  journal={Interdisciplinary perspectives on infectious diseases},
  year={2016},
  number={1}
}

@article{padi_valentine,
title = {PADI-web 3.0: A new framework for extracting and disseminating fine-grained information from the news for animal disease surveillance},
journal = {One Health},
volume = {13},
year = {2021},
author = {Sarah Valentin and Elena Arsevska and Julien Rabatel and Sylvain Falala and Alizé Mercier and Renaud Lancelot and Mathieu Roche}
}

@article{eventepi_abbood,
  title={EventEpi—A natural language processing framework for event-based surveillance},
  author={Abbood, Auss and Ullrich, Alexander and Busche, R{\"u}diger and Ghozzi, St{\'e}phane},
  journal={PLoS computational biology},
  volume={16},
  year={2020},
  publisher={Public Library of Science San Francisco, CA USA}
}

@article{event_extraction_survey,
  title={A survey of event extraction from text},
  author={Xiang, Wei and Wang, Bang},
  journal={IEEE Access},
  volume={7},
  year={2019}
}

@misc{liu2019roberta,
  author = {Liu, Yinhan and Ott, Myle and Goyal, Naman and Du, Jingfei and Joshi, Mandar and Chen, Danqi and Levy, Omer and Lewis, Mike and Zettlemoyer, Luke and Stoyanov, Veselin},
  note = {cite arxiv:1907.11692},
  title = {RoBERTa: A Robustly Optimized BERT Pretraining Approach},
  url = {http://arxiv.org/abs/1907.11692},
  year = 2019
}

@incollection{disease_surveillance_ml_survey_Zeng_2021,
  title={Artificial intelligence--enabled public health surveillance—from local detection to global epidemic monitoring and control},
  author={Zeng, Daniel and Cao, Zhidong and Neill, Daniel B},
  booktitle={Artificial intelligence in medicine},
  pages={437--453},
  year={2021},
  publisher={Elsevier}
}

@INPROCEEDINGS{disease_surveillance_ml_survey_Cabatuan_2020,
  author={Cabatuan, Melvin and Manguerra, Michael},
  booktitle={IEEE HNICEM}, 
  title={Machine learning for disease surveillance or outbreak monitoring: A review}, 
  year={2020}}

@inproceedings{reimers-2019-sentence-bert,
    title = "Sentence-BERT: Sentence Embeddings using Siamese BERT-Networks",
    author = "Reimers, Nils and Gurevych, Iryna",
    booktitle = "Proceedings of the 2019 Conference on Empirical Methods in Natural Language Processing",
    month = "11",
    year = "2019",
    publisher = "Association for Computational Linguistics",
    url = "http://arxiv.org/abs/1908.10084",
}

@article{Hubert1985,
  author    = {L. Hubert and P. Arabie},
  title     = {Comparing Partitions},
  journal   = {Journal of Classification},
  year      = {1985},
  volume    = {2},
  number    = {1},
  pages     = {193--218},
  doi       = {10.1007/BF01908075},
  publisher = {Springer}
}

@article{Strehl2002,
  author    = {Alexander Strehl and Joydeep Ghosh},
  title     = {Cluster Ensembles -- A Knowledge Reuse Framework for Combining Multiple Partitions},
  journal   = {Journal of Machine Learning Research},
  year      = {2002},
  volume    = {3},
  pages     = {583--617}
}

@inproceedings{Rosenberg2007,
  author    = {Andrew Rosenberg and Julia Hirschberg},
  title     = {V-Measure: A Conditional Entropy-Based External Cluster Evaluation Measure},
  booktitle = {Proceedings of the 2007 Joint Conference on Empirical Methods in Natural Language Processing and Computational Natural Language Learning (EMNLP-CoNLL)},
  year      = {2007},
  pages     = {410--420}
}

@misc{llm_fewshot,
      title={Large Language Models are Zero-Shot Reasoners}, 
      author={Takeshi Kojima and Shixiang Shane Gu and Machel Reid and Yutaka Matsuo and Yusuke Iwasawa},
      year={2023},
      eprint={2205.11916},
      archivePrefix={arXiv},
      primaryClass={cs.CL},
      url={https://arxiv.org/abs/2205.11916}, 
}

@article{DBLP:journals/corr/abs-1907-11692,
  author    = {Yinhan Liu and
               Myle Ott and
               Naman Goyal and
               Jingfei Du and
               Mandar Joshi and
               Danqi Chen and
               Omer Levy and
               Mike Lewis and
               Luke Zettlemoyer and
               Veselin Stoyanov},
  title     = {RoBERTa: {A} Robustly Optimized {BERT} Pretraining Approach},
  journal   = {CoRR},
  volume    = {abs/1907.11692},
  year      = {2019},
  url       = {http://arxiv.org/abs/1907.11692},
  archivePrefix = {arXiv},
  eprint    = {1907.11692},
  bibsource = {dblp computer science bibliography, https://dblp.org}
}

@article{Sanh2019DistilBERTAD,
  title={DistilBERT, a distilled version of BERT: smaller, faster, cheaper and lighter},
  author={Victor Sanh and Lysandre Debut and Julien Chaumond and Thomas Wolf},
  journal={ArXiv},
  year={2019},
  volume={abs/1910.01108}
}

@article{DBLP:journals/corr/abs-1909-11942,
  author    = {Zhenzhong Lan and
               Mingda Chen and
               Sebastian Goodman and
               Kevin Gimpel and
               Piyush Sharma and
               Radu Soricut},
  title     = {{ALBERT:} {A} Lite {BERT} for Self-supervised Learning of Language
               Representations},
  journal   = {CoRR},
  volume    = {abs/1909.11942},
  year      = {2019},
  url       = {http://arxiv.org/abs/1909.11942},
  archivePrefix = {arXiv},
  eprint    = {1909.11942},
  bibsource = {dblp computer science bibliography, https://dblp.org}
}

@article{DBLP:journals/corr/abs-1906-08237,
  author    = {Zhilin Yang and
               Zihang Dai and
               Yiming Yang and
               Jaime G. Carbonell and
               Ruslan Salakhutdinov and
               Quoc V. Le},
  title     = {XLNet: Generalized Autoregressive Pretraining for Language Understanding},
  journal   = {CoRR},
  volume    = {abs/1906.08237},
  year      = {2019},
  url       = {http://arxiv.org/abs/1906.08237},
  eprinttype = {arXiv},
  eprint    = {1906.08237},
  bibsource = {dblp computer science bibliography, https://dblp.org}
}

@article{DBLP:journals/corr/abs-1810-04805,
  author    = {Jacob Devlin and
               Ming{-}Wei Chang and
               Kenton Lee and
               Kristina Toutanova},
  title     = {{BERT:} Pre-training of Deep Bidirectional Transformers for Language
               Understanding},
  journal   = {CoRR},
  volume    = {abs/1810.04805},
  year      = {2018},
  url       = {http://arxiv.org/abs/1810.04805},
  archivePrefix = {arXiv},
  eprint    = {1810.04805},
  bibsource = {dblp computer science bibliography, https://dblp.org}
}

@article{DBLP:journals/corr/abs-1911-02116,
  author    = {Alexis Conneau and
               Kartikay Khandelwal and
               Naman Goyal and
               Vishrav Chaudhary and
               Guillaume Wenzek and
               Francisco Guzm{\'{a}}n and
               Edouard Grave and
               Myle Ott and
               Luke Zettlemoyer and
               Veselin Stoyanov},
  title     = {Unsupervised Cross-lingual Representation Learning at Scale},
  journal   = {CoRR},
  volume    = {abs/1911.02116},
  year      = {2019},
  url       = {http://arxiv.org/abs/1911.02116},
  eprinttype = {arXiv},
  eprint    = {1911.02116},
  bibsource = {dblp computer science bibliography, https://dblp.org}
}

@inproceedings{kakwani2020indicnlpsuite,
    title={{IndicNLPSuite: Monolingual Corpora, Evaluation Benchmarks and Pre-trained Multilingual Language Models for Indian Languages}},
    author={Divyanshu Kakwani and Anoop Kunchukuttan and Satish Golla and Gokul N.C. and Avik Bhattacharyya and Mitesh M. Khapra and Pratyush Kumar},
    year={2020},
    booktitle={Findings of EMNLP},
}

@inproceedings{event_extraction_2024_survey,
    title = "Generative Approaches to Event Extraction: Survey and Outlook",
    author = "Simon, {\'E}tienne  and
      Olsen, Helene  and
      You, Huiling  and
      Touileb, Samia  and
      {\O}vrelid, Lilja  and
      Velldal, Erik",
 
    booktitle = "Workshop on the Future of Event Detection (FuturED)",
   
    year = "2024"
}

@article{llm_chat_extract_materials,
  title={Extracting accurate materials data from research papers with conversational language models and prompt engineering},
  author={Polak, Maciej P and Morgan, Dane},
  journal={Nature Communications},
  volume={15},
  year={2024}
}

@misc{llm_chatie_2024,
      title={ChatIE: Zero-Shot Information Extraction via Chatting with ChatGPT}, 
      author={Xiang Wei and Xingyu Cui and Ning Cheng and Xiaobin Wang and Xin Zhang and Shen Huang and Pengjun Xie and Jinan Xu and Yufeng Chen and Meishan Zhang and Yong Jiang and Wenjuan Han},
      year={2024},
      eprint={2302.10205},
      archivePrefix={arXiv},
      primaryClass={cs.CL} 
}

@article{llm_gao_chatgpt_ee,
  title={Exploring the feasibility of chatgpt for event extraction},
  author={Gao, Jun and Zhao, Huan and Yu, Changlong and Xu, Ruifeng},
  journal={arXiv preprint arXiv:2303.03836},
  year={2023}
}

@misc{llama2,
      title={Llama 2: Open Foundation and Fine-Tuned Chat Models}, 
      author={Hugo Touvron and Louis Martin and Kevin Stone and Peter Albert and Amjad Almahairi and Yasmine Babaei and Nikolay Bashlykov and Soumya Batra and Prajjwal Bhargava and Shruti Bhosale and Dan Bikel and Lukas Blecher and Cristian Canton Ferrer and Moya Chen and Guillem Cucurull and David Esiobu and Jude Fernandes and Jeremy Fu and Wenyin Fu and Brian Fuller and Cynthia Gao and Vedanuj Goswami and Naman Goyal and Anthony Hartshorn and Saghar Hosseini and Rui Hou and Hakan Inan and Marcin Kardas and Viktor Kerkez and Madian Khabsa and Isabel Kloumann and Artem Korenev and Punit Singh Koura and Marie-Anne Lachaux and Thibaut Lavril and Jenya Lee and Diana Liskovich and Yinghai Lu and Yuning Mao and Xavier Martinet and Todor Mihaylov and Pushkar Mishra and Igor Molybog and Yixin Nie and Andrew Poulton and Jeremy Reizenstein and Rashi Rungta and Kalyan Saladi and Alan Schelten and Ruan Silva and Eric Michael Smith and Ranjan Subramanian and Xiaoqing Ellen Tan and Binh Tang and Ross Taylor and Adina Williams and Jian Xiang Kuan and Puxin Xu and Zheng Yan and Iliyan Zarov and Yuchen Zhang and Angela Fan and Melanie Kambadur and Sharan Narang and Aurelien Rodriguez and Robert Stojnic and Sergey Edunov and Thomas Scialom},
      year={2023},
      eprint={2307.09288},
      archivePrefix={arXiv},
      primaryClass={cs.CL} 
}

@inproceedings{nlp4pi_disease_map,
  title={From Text to Maps: LLM-Driven Extraction and Geotagging of Epidemiological Data},
  author={Harrod, Karlyn and Bhandari, Prabin and Anastasopoulos, Antonios},
  booktitle={ Workshop on NLP for Positive Impact},
  year={2024}
}

\appendix
\clearpage
\section{Technical Appendix}
\subsection{Building the Article Classifier}
\label{supp:art_cls_subsec}
A substantial portion of the articles (around 87\%) that pass the domain and the language filter are irrelevant to Health Sentinel. These articles can be from different genres of news articles including entertainment, crime, accidents, politics, finance--- none of which are related to health and must be discarded. Additionally, health news articles can be further divided into two types, Health Event  articles and Health Information articles. 
\begin{itemize}
    \item \textbf{Health Event: }These articles contain news regarding disease outbreaks, spread, and updates within specific geographical area. They are considered relevant to Health Sentinel as they contain actionable health events. Example: ``Two die of dengue in Mizoram, 1 in Manipur. Meanwhile forty-six cases of Chikungunya have been detected so far in Assam taking the total number of infections to 70''
    \item \textbf{Health Information: }These articles contain general information regarding different diseases, prevention methods and treatment options. They are considered irrelevant to Health Sentinel as they do not contain any actionable events. Example: ``What is Dengue? How does dengue spread and 10 ways to stay safe this monsoon''
\end{itemize}
Given that this stage of the pipeline encounters a massive amount of articles that belong to 13 different languages(English + 12 Indic languages), we developed separate binary classifiers for each supported language. We intentionally avoided translating all articles to English before classification, as this would be is a time-consuming process that compromises our ability to maintain low latency. Considering that over 87\% of articles are anyway irrelevant to our system at this stage, this step did not have any payoff. 

To train these classifiers, we collected 34,527 English articles from different genres to ensure broad representation. The positive class comprised of 7,374 articles covering all diseases that were considered as important and additional relevant diseases. Special care was taken to include health information articles about the same diseases in the negative class, which contained 27,153 articles. Additionally, we collected a wide range of non-health related news genres as part of the negative class to ensure classification robustness. These articles were all then translated to the 12 different supported Indic languages using \texttt{IndicTrans2} model to create 13 separate datasets for fine tuning each model.

For selecting the best model, we selected 6 different pretrained models for each of the 13 supported languages as follows:
\begin{itemize}
    \item \textbf{English:}\newline \texttt{roberta-base}~\cite{DBLP:journals/corr/abs-1907-11692}, \texttt{distilbert-base-cased}~\cite{Sanh2019DistilBERTAD}, \texttt{albert-base-v2}~\cite{DBLP:journals/corr/abs-1909-11942}, \texttt{xlnet-base-cased}~\cite{DBLP:journals/corr/abs-1906-08237}, \texttt{bert-base-cased}~\cite{DBLP:journals/corr/abs-1810-04805}, \texttt{bert-base-uncased}~\cite{DBLP:journals/corr/abs-1810-04805}.
    
    \item \textbf{12 Indic Languages:}\newline \texttt{xlm-roberta-base}~\cite{DBLP:journals/corr/abs-1911-02116}, \texttt{google/muril-base-cased}~\cite{khanuja2021murilmultilingualrepresentationsindian}, \texttt{ai4bharat/indic-bert}~\cite{kakwani2020indicnlpsuite}, \texttt{bert-base-multilingual-cased}~\cite{DBLP:journals/corr/abs-1810-04805}.
\end{itemize}

All models were finetuned separately for each of the 13 supported languages and the best performing models were selected based on recall. Recall was chosen as the primary metric as the article classifier functions as a soft filter to remove majority of the irrelevant articles. Allowing some of the irrelevant articles to pass through is not a major concern, but it is essential to maximize the number of relevant articles retained at this stage. The best performing models and their metric scores are shown in Table \ref{tab:main_eval_artclf}.



\subsection{Event Extraction}
In this section we provide the details on questions and hypothesis formulation for QA and NLI based pipeline, as well as prompt design for LLM based pipeline. 
\subsubsection{Question formulation for the QA model}
The QA model is used to extract numbered events from the articles. We use the templates shown in Table \ref{tab:questions} to formulate questions based on previously extracted entities: disease and location. The questions contain the combination of entities: Incident and Incident type, while the remaining entity Number is extracted by the QA model. If the model returns similar values of number across different question categories, the one with the highest confidence is considered.  
\begin{table*}[!hbt]
 \centering
    \renewcommand{\arraystretch}{1.3}
    \setlength{\tabcolsep}{10pt}
    \rowcolors{2}{gray!15}{white} 
    \begin{tabular}{>{\centering\arraybackslash}m{0.15\textwidth} m{0.75\textwidth}} 
        \toprule
        \rowcolor{blue!20} \textbf{Category} & \textbf{Questions} \\
        \midrule
        \textnormal{new\_cases} & 
        \begin{itemize}
            \setlength{\itemsep}{0pt}
            \item How many new DISEASE cases were reported in LOCATION?
            \item How many new DISEASE cases were reported in LOCATION in the last 24 hours?
            \item How many fresh DISEASE cases were reported in LOCATION?
            \item How many fresh DISEASE cases were reported in LOCATION in the last 24 hours?
            \item How many new DISEASE infections were reported in LOCATION?
            \item How many fresh DISEASE infections were reported in LOCATION?
            \item How many DISEASE cases were reported in LOCATION in 24 hours?
        \end{itemize} \\
        \midrule
        \textnormal{new\_deaths} & 
        \begin{itemize}
            \setlength{\itemsep}{0pt}
            \item How many new DISEASE deaths were reported in LOCATION?
            \item How many new DISEASE deaths were reported in LOCATION in the last 24 hours?
            \item How many fresh DISEASE deaths were reported in LOCATION?
            \item How many fresh DISEASE deaths were reported in LOCATION in the last 24 hours?
            \item How many new deaths due to DISEASE were reported in LOCATION?
            \item How many DISEASE deaths were reported in LOCATION in 24 hours?
        \end{itemize} \\
        \midrule
        \textnormal{total\_cases} & 
        \begin{itemize}
            \setlength{\itemsep}{0pt}
            \item How many total DISEASE cases were reported in LOCATION?
            \item What is the total number of DISEASE cases reported in LOCATION?
            \item How many total cases of DISEASE were reported in LOCATION?
            \item What is the total tally of DISEASE cases reported in LOCATION?
        \end{itemize} \\
        \midrule
        \textnormal{total\_deaths} & 
        \begin{itemize}
            \setlength{\itemsep}{0pt}
            \item How many total DISEASE deaths were reported in LOCATION?
            \item How many total deaths due to DISEASE were reported in LOCATION?
            \item What is the total number of deaths due to DISEASE in LOCATION?
            \item How many total deaths of DISEASE were reported in LOCATION?
            \item What is the total tally of DISEASE deaths in LOCATION?
        \end{itemize} \\
        \bottomrule
    \end{tabular}
    \caption{Question templates for different combinations of the entities `Incident type' and `Incident'. The disease and location values extracted in earlier stages of the pipeline are inserted into these templates to generate specific questions.}
    \label{tab:questions}
\end{table*}

\subsubsection{Hypothesis formulation for the NLI model}
Articles where the QA model does not extract any events, are processed by the NLI model for numberless event extraction. In a manner similar to the questions, we construct hypotheses showcasing the presence of relevant health events, which are then validated by the NLI model with the article serving as premise. The hypothesis templates are shown in Table \ref{tab:hypothesis}. If a hypothesis is validated with an entailment score greater than 0.5, the corresponding event is generated. 
\begin{table*}[!hbt]
    \centering
    \renewcommand{\arraystretch}{1.3}
    \setlength{\tabcolsep}{10pt}
    \rowcolors{2}{gray!15}{white} 
    \begin{tabular}{>{\centering\arraybackslash}m{0.15\textwidth}m{0.75\textwidth}}
        \toprule
        \rowcolor{blue!20} \textbf{Category} & \textbf{Hypotheses} \\
        \midrule
        \textnormal{cases} & 
        \begin{itemize}
            \setlength{\itemsep}{0pt}
            \item DISEASE is spreading in LOCATION
            \item DISEASE was spreading in LOCATION
            \item DISEASE has been spreading in LOCATION
            \item Cases of DISEASE increased in LOCATION
            \item Cases of DISEASE are increasing in LOCATION
            \item Cases of DISEASE have risen in LOCATION
            \item Cases of DISEASE are rising in LOCATION
            \item A person is infected by DISEASE in LOCATION
            \item A person was infected by DISEASE in LOCATION
            \item A person was diagnosed with DISEASE in LOCATION
            \item A person was affected by DISEASE in LOCATION
            \item People are infected by DISEASE in LOCATION
            \item People were infected by DISEASE in LOCATION
            \item People are suffering from DISEASE in LOCATION
            \item People are sick with DISEASE in LOCATION
            \item A DISEASE outbreak was reported in LOCATION
        \end{itemize} \\
        \midrule
        \textnormal{deaths} & 
        \begin{itemize}
            \setlength{\itemsep}{0pt}
            \item People died due to DISEASE in LOCATION
            \item Deaths were reported in LOCATION due to DISEASE
            \item Deaths are reported in LOCATION due to DISEASE
            \item People are dying of DISEASE in LOCATION
            \item Deaths have been reported in LOCATION due to DISEASE
            \item Deaths have occurred due to DISEASE in LOCATION
        \end{itemize} \\
        \bottomrule
    \end{tabular}
    \caption{Hypothesis templates for different event categories (`cases' and `deaths`). The table presents various templates used to generate hypotheses based on the extracted values of disease and location. Given an article as the premise, these hypotheses are validated by the NLI model. If the hypothesis is entailed, the corresponding event is generated.}
    \label{tab:hypothesis}
\end{table*}

\subsubsection{Prompt Designing for LLM}
For the task of event extraction from articles, precise and well-structured prompts are essential for obtaining reliable outputs from LLMs. In our experiments, we tested various prompts to identify the most effective one. One such prompt is shown in Table \ref{tab:llm_prompt}. Following the guidelines on prompt engineering by OpenAI\footnote{\url{https://platform.openai.com/docs/guides/prompt-engineering}}, the LLM is assigned the persona of an Event Extractor, to align its output with the specific requirements of our task.

\begin{table*}[!hbt]
    \centering
    \begin{tabular}{|p{0.96\textwidth}|} 
        \hline
        \multicolumn{1}{|c|}{\textbf{System Prompt}} \\
        \hline
        You are a renowned event extractor specializing in identifying disease outbreaks within Indian news articles. Your expertise lies in meticulously pinpointing health events with high accuracy. Your task is to analyze an English article as input, carefully extract health events, and provide them in a structured format. A health event is an unusual occurrence in a specific area that could potentially threaten the health of people. This includes:  
        
        \textbf{Unusual sickness in people:} This could be one case of a rare disease, a sudden increase in cases of a common illness, or people falling ill due to some unidentified reason.  
        
        \textbf{Animals getting sick:} If there's a sudden jump in animals getting a particular disease, especially one that can spread to people, it's a health event.  
        
        \textbf{Animal Bites:} Incidents of animals biting humans in a specific area.  
        
        These health events serve as signals for public health officials to promptly investigate and take necessary actions to safeguard public health.  
        
        Note that it is critical to differentiate between health events and health information. Health information consists of a broad spectrum of info related to human health and well-being. This includes: disease prevention, medical research, public health initiatives, guidelines, and action plans developed to combat infectious diseases.
        
        The extracted event should be presented in the following JSON schema:  
        [
        \{`Disease': `The name of the disease mentioned in the article',  
        `Location': `The most local geographical level affected (e.g., state, city, or district)',  
        `Incident (case or death)': `Indicates whether the article discusses cases of illness or deaths',  
        `Incident Type (new or total)': `Specifies if the article refers to new or total cases/deaths',  
        `Number': `The numerical value associated with the incident'\}]
        
        STRICTLY focus on events occurring in India, disregarding any news outside the country.  
        \\
        \hline
    \end{tabular}

    \vspace{0.4cm}

    \begin{tabular}{|p{0.468\textwidth}|p{0.468\textwidth}|} 
        \hline
        \multicolumn{2}{|c|}{\textbf{Few-Shot Examples}} \\
        \hline
        \textbf{Input Article} & \textbf{Output} \\ 
        \hline
        \textbf{E1:} Ambikapur News: Four people of the same family fell ill after eating putu. Four members of the same family fell ill after consuming wild puttu on Sunday night in Parpatia village of Mainpat development block of Chhattisgarh.  
        & 
        [ \{`Disease': `ill after consuming food', `Location': `Ambikapur', `Incident (case or death)': `case', `Incident Type (new or total)': `new', `Number': `4'\} ] \\
        \hline
        \textbf{E2:} 8 laborers died when the truck overturned. Bihar Accident: 8 people died when a truck carrying a load of pipes overturned. Some others were seriously injured. This incident happened in Purnia, Bihar.  
        & 
        \texttt{[]} \\
        \hline
        \textbf{E3:} 3,353 vaccinated against rabies in government hospitals. Coimbatore: During the current year, 2,539 rabies cases and 814 cases of dog bites have been reported in Government Hospitals.  
        & 
        \texttt{[]} \\
        \hline
        \textbf{E4:} 906 new cases of Covid-19 were reported in India, the number of patients under treatment decreased to 10,179. India In Hindi | According to the updated data released by the Union Health Ministry at eight o'clock on Thursday morning, after the death of 20 more patients from Covid-19, the total number of people who lost their lives due to coronavirus infection in the country has increased to 5,31,814.  
        & 
        [
        \{`Disease': `Corona', `Location': `India', `Incident (case or death)': `case', `Incident Type (new or total)': `new', `Number': `906'\}, 
        \{`Disease': `Corona', `Location': `India', `Incident (case or death)': `death', `Incident Type (new or total)': `new', `Number': `20'\}, 
        \{`Disease': `Corona', `Location': `India', `Incident (case or death)': `death', `Incident Type (new or total)': `total', `Number': `531814'\} 
        ] \\
        \hline
    \end{tabular}
    
    \caption{System prompt with few-shot examples for extracting health events from news articles.}
    \label{tab:llm_prompt}
\end{table*}


\begin{table*}[!hbt]
\centering
\footnotesize
\begin{tabular}{lcccccc}
\toprule
\textbf{Model} & \textbf{Precision} & \textbf{Recall} & \textbf{F1} & \textbf{Exact Match} & \textbf{Detection Rate} \\
\midrule
GPT-4o-Mini (Zero-Shot) & 0.60 & 0.58 & 0.59 & 0.52 & 0.91 \\
GPT-4o-Mini (Few-Shot) & \textbf{0.70} & \textbf{0.67} & \textbf{0.68} & \textbf{0.61} & \textbf{0.92} \\
\bottomrule
\end{tabular}
\caption{Comparison of GPT-4o-Mini performance for zero-shot and few-shot prompt. The few-shot prompt include four examples from Table \ref{tab:llm_prompt}, leading to significant improvement in precision, recall, and exact match over zero-shot method.}
\label{tab:gpt4o_comparison}
\end{table*}

The prompt is carefully designed to focus on extracting infectious disease outbreaks from Indian news articles. It provides clear definition of each entity present in the event list and explicitly differentiates between relevant health events and general health information. Additionally, the LLM is instructed to exclude events related to international locations.

To guide the extraction process, the prompt is supplemented with few-shot examples. These examples include both relevant articles with actionable events and irrelevant general health information,  helping LLM to distinguish between relevant and irrelevant content, enhancing the overall accuracy of event extraction. To showcase the effectiveness of few-shot approach, we compare it with zero-shot approach in Table \ref{tab:gpt4o_comparison}. Few-shot examples lead to a significant improvement in both precision and recall.

\begin{figure*}[!hbt]
\centering
\includegraphics[width=1.0\textwidth]{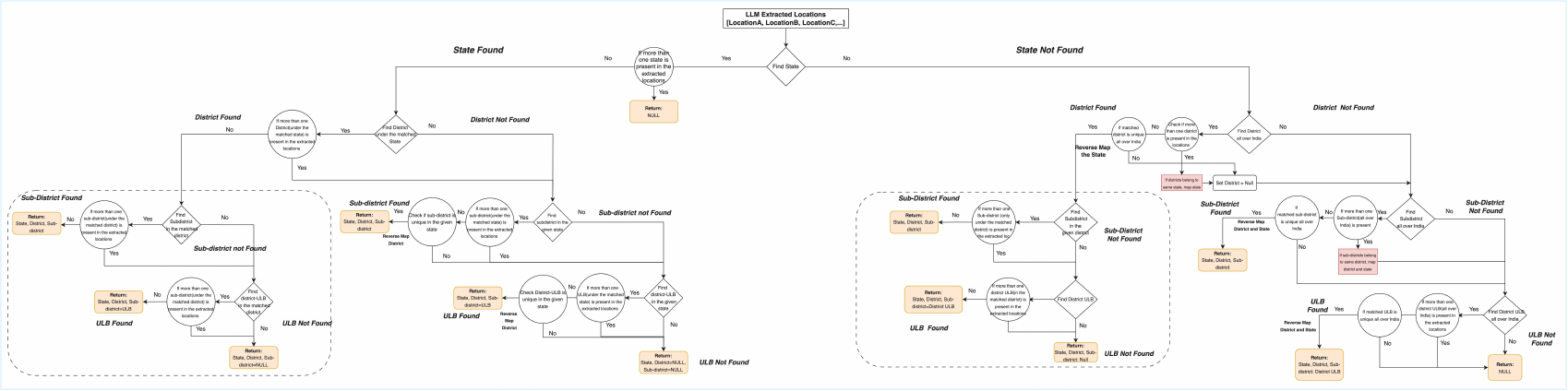}
\caption{Logic for mapping extracted locations to appropriate State, District, Sub-district, and Urban Local Bodies (ULBs). First, individual locations are extracted from the comma separated values. The process starts with assigning a state if present, followed by assigning a district and sub-district/ULB. If a state is not identified, the logic tries to assign a district or sub-district/ULB and then reverse maps to determine the corresponding state. If multiple values are found during assignment, the location is not mapped.}
\label{fig:location_mapping}
\end{figure*}

\subsection{Mapping of Disease and Location}
\label{sec:disease_mapping}
\subsubsection{Disease Mapping}
In the extracted events, diseases are often present in colloquial or media-specific terms rather than the standardized nomenclature used by health authorities. For example, disease `Pneumonia' is sometimes referred to as `Lung Fever'.
Additionally, disease extracted using LLMs may include extra words with the disease name, such as `Cholera Infection' or `Cholera Infectious Disease' instead of just `Cholera'. Since our solution is tailored for use by health authorities, it's important to map these terms to the standardized disease names.
The mapping process is performed in two stages:

\begin{table*}[!ht]
    \centering
    \begin{tabular}{|p{0.90\textwidth}|}
        \hline
        \multicolumn{1}{|c|}{\textbf{System Prompt}} \\
        \hline
        As a renowned disease mapper, you are tasked with mapping a given disease to the nearest standard disease name in the provided list. You should only map diseases where you are certain of a close similarity, otherwise, label it as 'Others'. Your disease list is $\{Disease List\}$. \\
        \hline
        \multicolumn{1}{|c|}{\textbf{Few-Shot Examples}} \\
        \hline
        \textbf{Input:} "sick after eating contaminated food" \\
        \textbf{Output:} "Food Poisoning infection" \\
        \hline
        \textbf{Input:} "Diarrhoea outbreak" \\
        \textbf{Output:} "Acute Diarrhoeal Disease" \\
        \hline
        \textbf{Input:} "Bird flu (H5N1)" \\
        \textbf{Output:} "Bird flu" \\
        \hline
        \textbf{Input:} "Cricket Fever" \\
        \textbf{Output:} "Others" \\
        \hline
    \end{tabular}
    \caption{System prompt and few-shot examples for mapping diseases to standard names using an LLM.}
    \label{tab:disease_prompt}
\end{table*}

\begin{enumerate}
    \item \textbf{Synonym Mapping}: To standardize the name of extracted diseases, we use a dictionary that contains a mapping of common disease synonyms and media terms to standard names. This dictionary is curated and verified by public health experts.
    \item \textbf{Synonym Expansion Using LLM:} To make the synonym dictionary comprehensive, we prompt an LLM to map the un-mapped diseases from the previous stage to the nearest name in the list of 122 standard diseases within reason. Any new synonym identified is added to our dictionary, following verification by experts. If an appropriate mapping is not present from the 122 standard diseases, the disease is mapped to a miscellaneous category called "Others". The prompt used for this mapping can be found in Table \ref{tab:disease_prompt}.
\end{enumerate}

\subsubsection{Location Mapping}
The extracted location data may include information such as the names of villages, districts, and states. However, it is essential to map these to the appropriate administrative levels, such as State, District, or Sub-district, so that the relevant health authorities can be prompted to take action. Similar to disease mapping, this process is performed in two parts:


\begin{enumerate}
    \item \textbf{Logic-Based Mapping:} We use a standard hierarchical dictionary of \texttt{States $\rightarrow$ Districts $\rightarrow$ Sub-districts $\rightarrow$ Urban Local Bodies (ULBs)} and their synonyms to assign appropriate values to the extracted location data. The logic first assigns each individual location to a state, then to a district, and finally to a sub-district or ULB. Additionally, backward mapping--- such as from district to state or sub-district to district is performed to handle cases where where direct mapping is not possible. A visual representation of this mapping process is show in Figure \ref{fig:location_mapping}.

    \item \textbf{LLM-Based Mapping:} For locations that cannot be mapped using the logic-based approach, we use an LLM to assist in identifying the correct administrative levels. The LLM is tasked with extracting the Indian state and district from the given article, identifying any international locations, or returning an empty result if the location cannot be mapped.
    Due to the extensive knowledge embedded within LLMs, they often perform accurate state-level mapping. However, it is prone to hallucinations in case of district mapping. To mitigate this, we prompt the LLM to perform the mapping multiple times and only accept it if all the outputs are consistent. The prompt used can be seen in Table \ref{tab:location_prompt}. If an appropriate mapping cannot be found, the corresponding fields are left blank (`` '').
\end{enumerate}
\begin{table*}[h!]
    \centering
    \begin{tabular}{|p{0.90\textwidth}|}
        \hline
        \multicolumn{1}{|c|}{\textbf{System Prompt}} \\
        \hline
        You are an expert in extracting locations of occurrence of health events, with a capability of distinguishing Indian locations with international, and providing precise details down to state and district levels within India. Your task is to analyze the provided English article and identify the event's location and classify it accordingly. \\
        1. For Indian events: Return the state and district (only if mentioned in the article). \\
        2. If the event relates to India but cannot be pinpointed to a specific state or district, return blank values for state and district. \\
        3. For international events: Return the output as `International'. \\
        4. For events discussing locations in both India and outside India: Return a blank value for the state and district. \\
        \hline
        \multicolumn{1}{|c|}{\textbf{Few-Shot Examples}} \\
        \hline
        \textbf{Input:} `Four people of the same family fell ill after eating putu. Four members of the same family fell ill after consuming wild puttu on Sunday night in Parpatia village of Mainpat development block of Chhattisgarh.' \\
        \textbf{Output:} [\{\{`State': `Chhattisgarh', `District': `Surguja'\}\}] \\
        \hline
        \textbf{Input:} `906 new cases of Malaria were reported from Gaya, the number of patients' deaths has reached 50. Bihar In Hindi.' \\
        \textbf{Output:} [\{\{`State': `Bihar', `District': `Gaya, Darbhanga'\}\}] \\
        \hline
        \textbf{Input:} `Bird flu hits Northwest Iowa dairies - Storm Lake Times Pilot.' \\
        \textbf{Output:} [\{\{`State': `International', `District': ``\}\}] \\
        \hline
        \textbf{Input:} `Signs of bird flu in 4 states - Government of India Signs of bird flu in 4 states.' \\
        \textbf{Output:} [\{\{`State': `', `District': `'\}\}] \\
        \hline
    \end{tabular}
    \caption{System prompt and few-shot examples for extracting state and district.}
    \label{tab:location_prompt}
\end{table*}

\subsection{Clustering}
\label{sec:clustering}
Once all events are extracted and processed, they are clustered together to perform de-duplication. This enables us to deal with multiple media outlets covering the same occurrence of an event and isolating all the unique health events. We achieve this by combining an ML based approach with a finely curated set of rules. Clustering is performed day-wise to categorize unique events based on their occurrence date.

\subsubsection{Methodology}
\begin{figure*}[!hbt]
\centering
\includegraphics[width=1.0\textwidth]{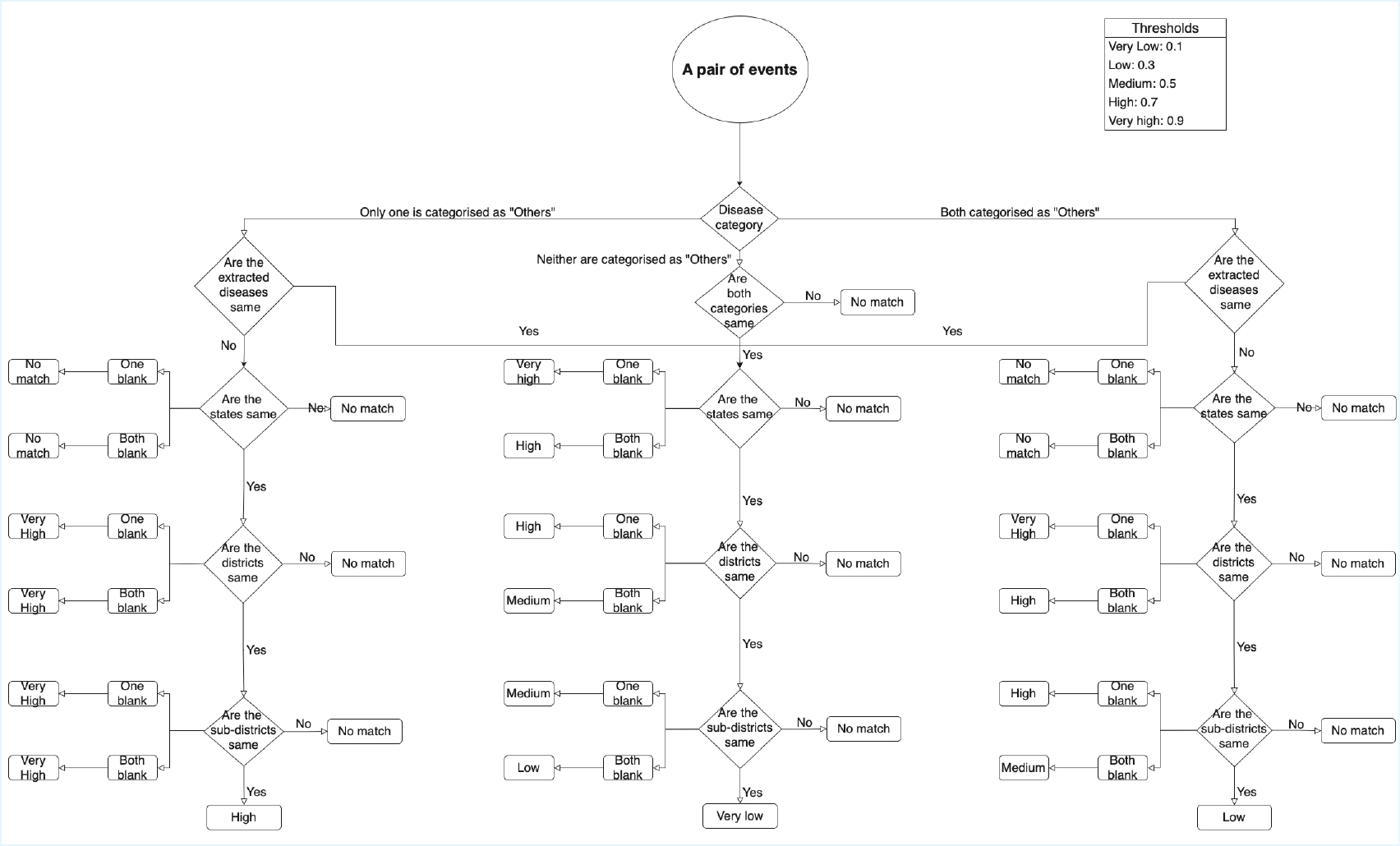}
\caption{Logic flow of the rules that are used to determine the threshold that is applied on the similarity score for a pair of events}
\label{fig:clustering_flow}
\end{figure*}

The following steps are undertaken to create clusters with the ideal compositions and thus identify all unique events on an daily basis:
\begin{itemize}
    \item All the events extracted for the present day are collected.
    \item Every event has an associated article from which it was extracted. We use a sentence transformer, \emph{paraphrase-distilroberta-base-v2}~\cite{reimers-2019-sentence-bert} to generate embeddings for all the articles associated with each of the extracted events.
    \item We use cosine similarity to compute pairwise similarity scores for each events' associated articles. An example of how this matrix looks like can be seen in Table \ref{tab:example_simimat2D}.
    \item Following this, a rule based approach is taken to fix the threshold that needs to be applied for each pair of events to determine if they are a match. The logical flow of the rules used for determining these thresholds are shown in Figure \ref{fig:clustering_flow}. 
    \item After the thresholds are determined based on the rules and applied to the similarity score of each pair of events, we get a match matrix with 1's an 0's. An example of how this matrix looks like can be seen in Table \ref{tab:example_matchmat2D}.
    \item We then treat the match matrix as a graph problem, where a 1 represents the presence of an edge between a pair of events and a 0 represents otherwise. We use a Depth First Search approach to identify all the disjoint graphs from the match matrix and treat each of them as a cluster. For the given example in Table \ref{tab:example_matchmat2D}, we get two clusters as follows:
    \begin{itemize}
        \item Cluster A: Event 1, Event 3, and Event 5.
        \item Cluster B: Event 2, and Event 4.
    \end{itemize}
    While cluster B's formation is straightforward, it must be noted that even though Event 1 and Event 3 are not matched, they can end up in the same cluster as they are chained through Event 5.
    \item This chaining effect is usually observed due to the presence of events with ambiguous information. This ambiguity can occur in two different ways as follows:
    \begin{itemize}
        \item Disease ambiguity: This phenomenon is encountered when an event's extracted disease is mapped to "Others".
        \item Location ambiguity: This phenomenon is encountered when the mapped state, district, or sub district is blank(``'').
    \end{itemize}
    This sometimes leads to the formation of clusters that have events with conflicting information and are chained through an ambiguous event. Example: An event with a district as ``Mallapuram'' is clustered with another event with district as ``Kozhikode'' due to the presence of an ambiguous event with a blank district(``'').
    \item We thus have an additional step to detect clusters with conflicting information. On detection, they are further broken down into multiple clusters without any conflicting information.
\end{itemize}

\begin{table*}[t]
\centering
\begin{tabular}{lccccc}
\toprule
 & \textbf{Event 1} & \textbf{Event 2} & \textbf{Event 3} & \textbf{Event 4} & \textbf{Event 5} \\
\midrule
\textbf{Event 1} & 1.0 & 0.54 & 0.23 & 0.48 & 0.75 \\
\textbf{Event 2} & 0.54 & 1.0 & 0.43 & 0.84 & 0.16 \\
\textbf{Event 3} & 0.23 & 0.43 & 1.0 & 0.38 & 0.89 \\
\textbf{Event 4} & 0.48 & 0.84 & 0.38 & 1.0 & 0.73 \\
\textbf{Event 5} & 0.75 & 0.16 & 0.89 & 0.73 & 1.0 \\
\bottomrule
\end{tabular}
\caption{Example of a 2D matrix created by using a sentence transformer followed by cosine similarity computation for 5 events}
\label{tab:example_simimat2D}
\end{table*}

\begin{table*}[t]
\centering
\begin{tabular}{lccccc}
\toprule
 & \textbf{Event 1} & \textbf{Event 2} & \textbf{Event 3} & \textbf{Event 4} & \textbf{Event 5} \\
\midrule
\textbf{Event 1} & 1 & 0 & 0 & 0 & 1 \\
\textbf{Event 2} & 0 & 1 & 0 & 1 & 0 \\
\textbf{Event 3} & 0 & 0 & 1 & 0 & 1 \\
\textbf{Event 4} & 0 & 1 & 0 & 1 & 0 \\
\textbf{Event 5} & 1 & 0 & 1 & 0 & 1 \\
\bottomrule
\end{tabular}
\caption{Example of a 2D matrix created after applying the corresponding thresholds for each pair of events}
\label{tab:example_matchmat2D}
\end{table*}

    

\subsubsection{Evaluation Metrics}
\label{clustering_metrics}
To quantitatively evaluate the quality of the clusters formed, we employ three key metrics: the Adjusted Rand Index (ARI)~\cite{Hubert1985}, Normalized Mutual Information (NMI)~\cite{Strehl2002}, and V-Measure~\cite{Rosenberg2007}. These metrics are calculated on a per day basis, allowing for a detailed and dynamic assessment of clustering performance over the different dates in our clustering dataset, as illustrated in Table \ref{tab:combined_cluster}.

\noindent\textbf{Adjusted Rand Index (ARI):}
The ARI provides a normalized measure of the similarity between two data clustering, after correcting agreements occurring by random chance. It is particularly useful in determining the agreement between the ground truth labels and the clusters generated by our algorithm. The ARI is calculated as follows:
\[
ARI = \frac{RI - \text{Expected}[RI]}{\text{Max}[RI] - \text{Expected}[RI]}
\]
where $RI$ (Rand Index) measures the agreement of the clustering with the true labels, defined by:
\[
RI = \frac{TP + TN}{TP + FP + FN + TN}
\]
Here, $TP$ (true positives) and $TN$ (true negatives) are pairs correctly identified as belonging to the same or different clusters. While $FP$ (false positives) and $FN$ (false negatives) are pairs incorrectly identified as belonging together or apart.

\noindent\textbf{Normalized Mutual Information (NMI):}
NMI is an adjustment of the Mutual Information (MI) score that accounts for the chance grouping of elements, normalized by the entropy of the clusters. This makes it a reliable metric for comparing clustering of different sizes and compositions. It is computed as:
\[
NMI = \frac{2 \times I(y; \hat{y})}{H(y) + H(\hat{y})}
\]
where $I(y; \hat{y})$ represents the mutual information between the predicted and true labels, indicating the amount of information gained about one through the other. $H(y)$ and $H(\hat{y})$ are the entropy of the true labels and the predicted labels, respectively.

\noindent\textbf{V-Measure:}
This metric offers a balance between homogeneity (each cluster contains only members of a single class) and completeness (all members of a given class are assigned to the same cluster). The V-Measure is defined as the harmonic mean of these two aspects, providing a single score to assess the effectiveness of clustering without the need for each cluster to be of approximately equal size:
\[
V\text{-}Measure = \frac{\text{Homogeneity} \times \text{Completeness}}{\text{Homogeneity} + \text{Completeness}}
\]
where Homogeneity and Completeness are calculated based on the distribution of each class within the clusters and the consistency of class labels within each cluster.

\subsection{Hardware and Software Configuration}
The pipeline presented in this paper runs on a machine with Ubuntu 20.04.6 LTS operating system, with an Intel(R) Xeon(R) CPU @ 2.00GHz and an NVIDIA T4 GPU with 16 GB of GPU RAM. The pipeline is implemented in Python 3.9. We used an OpenAI API\footnote{\url{https://openai.com/index/openai-api/}} for proprietary LLMs such as GPT-3.5-Turbo and GPT-4o-mini. For open source LLMs, we used the latest instruction fine-tuned 4-bit quantized versions of models: \texttt{llama3.1-8b} and \texttt{gemma2-9b} provided by Ollama\footnote{\url{https://ollama.com/}}. The training of certain models was performed on 32GB V100 GPUs, while all the inferences and evaluation were performed on a 16GB NVIDIA T4 GPU. 

\end{document}